\documentclass[journal]{IEEEtran}

\ifCLASSINFOpdf
\else
\fi
\usepackage{graphicx}
\usepackage{booktabs}
\usepackage{amsmath}
\usepackage{rotating}
\usepackage[pagebackref=true,breaklinks=true,colorlinks,linkcolor=blue,citecolor=blue]{hyperref}
\usepackage{color}
\usepackage{times}
\usepackage{epsfig}
\usepackage{graphicx}
\usepackage{amssymb}
\usepackage{multirow}
\usepackage{algorithm}
\usepackage{algorithmic}
\usepackage{array}
\usepackage{cite}
\newcommand{\RN}[1]{%
  \textup{\uppercase\expandafter{\romannumeral#1}}%
}

\usepackage[utf8]{inputenc}
\usepackage[T1]{fontenc}
\usepackage{textcomp}
\usepackage[table]{xcolor}
\usepackage{threeparttable}
\usepackage{makecell} 
\usepackage{dcolumn}
\usepackage{siunitx}
\usepackage{pifont}
\newcommand{\xmark}{\ding{55}}
\newcommand{\ie}{\emph{i.e.}}
\newcommand{\eg}{\emph{e.g.}}

\usepackage{cleveref}
\begin{document}

\title{Disentangled Fine-Grained Prototype Learning for Incomplete Image-Tabular Classification}
\author{Feixiang Zhou, Jianyang Xie, Zhuangzhi Gao, Qinkai Yu, Fu Wang, Yuheng Fan, Jing Li, Zheheng Jiang, Yitian Zhao, \textit{Senior Member, IEEE}, Yanda Meng, He Zhao, Gregory Y.H. Lip, and Yalin Zheng
\thanks{F. Zhou, J. Xie, Z. Gao, F. Wang, Y. Fang, H. Zhao, and Y. Zheng are with the School of Eye and Vision Sciences, University of Liverpool, U.K. Y. Zheng is the corresponding author.}
\thanks{G. Y.H. Lip is with the Department of Cardiovascular and Metabolic Medicine, University of Liverpool, U.K.}
\thanks{Q. Yu is with the School of Computer Science, University of Exeter, U.K.}
\thanks{J. Li is with the School of Computer Science and Engineering, South China University of Technology, China.}
\thanks{Z. Jiang is with the School of Computing and Mathematical Sciences, University of Leicester, U.K.}
\thanks{Y. Zhao is with Ningbo Institute of Industrial Technology, Chinese Academy of Sciences, China.}
\thanks{Y. Meng is with the Bioengineering Program, Biological and Environmental Science and Engineering Division (BESE), King Abdullah University of Science and Technology (KAUST), Saudi Arabia.}

}

\maketitle

\begin{abstract}
The missing-modality problem poses a significant challenge in image–tabular multimodal learning across a wide range of multimedia applications, including product understanding, recommendation systems, and medical diagnosis. This challenge is particularly pronounced when the two modalities are highly heterogeneous, as images and tabular attributes differ substantially in their semantic granularity and data distributions.  Existing methods learn modality-invariant representations through disentanglement and alignment over global token-averaged features, capturing only coarse cross-modal consistency and overlooking fine-grained semantic and distributional misalignment, which hampers the exploitation of complementary cues under missing modalities. To address this, we propose DFPL, a novel framework for fine-grained prototype learning. Specifically, Shared–Specific Prototype Modeling (SSPM) extracts compact and diverse shared and modality-specific prototypes, and further performs prototype-level disentanglement to suppress redundant intra-modality correlations. Additionally, we propose a Prototype-guided Fine-grained Alignment (PFA) module that jointly enforces prototype-level distribution matching and prototype-to-class semantic alignment within a unified prototype space, thereby preserving both fine-grained distributional and semantic consistency across modalities. We further introduce a Class-aware Multi-scale Aggregation (CMA) module to adaptively aggregate shared semantics and modality-specific characteristics from global and prototype levels for robust predictions. Extensive experiments on three diverse image–tabular benchmarks demonstrate the superiority of our method compared to the previous approaches under various missing-modality settings. Code will be made publicly available.
\end{abstract}


\section{Introduction}
\label{sec:intro}
\begin{figure}[t]
  \centering
   \includegraphics[width=1\linewidth]{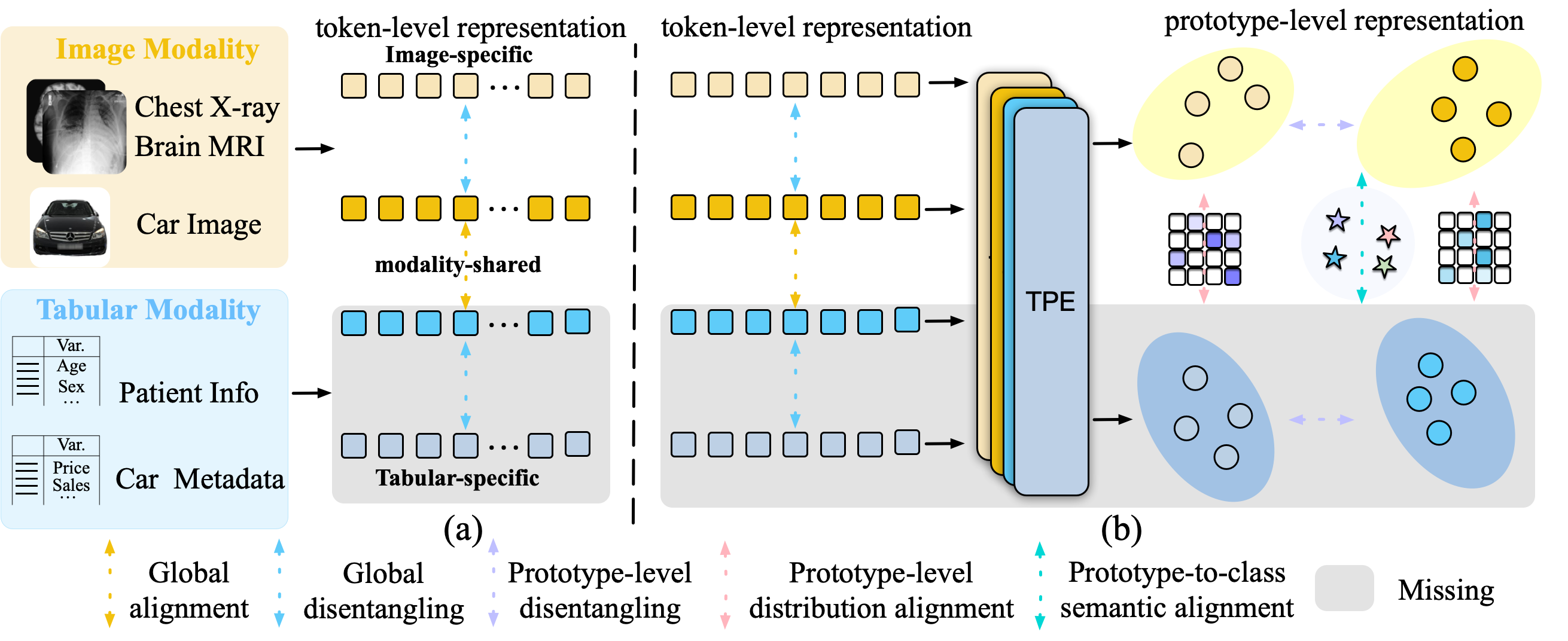}
   \caption{(a) Most existing shared-specific modeling methods \cite{du2025stil,yao2024drfuse} perform disentanglement and alignment only on globally averaged features,  leading to suboptimal intra-modality disentanglement and inter-modality alignment. (b) Unlike the coarse-grained modeling in (a), our method further employs a Token Prototype Extractor (TPE) to derive both shared and modality-specific prototypes, followed by prototype-level disentanglement and alignment. This enables more effective use of fine-grained semantic and distributional cues from the available modality to offset the missing one.
   }
   \label{fig:challenge}
\end{figure}

\IEEEPARstart{M}{U}{L}{T}{I}{M}{O}{D}{A}{L} learning has achieved remarkable progress in a wide range of applications, including vision–language understanding \cite{zhang2024vision,yan2025multi}, audio–visual analysis \cite{zhuang2025intra,jiang2025boosting}, and medical diagnosis \cite{dong2025multimodal,qiu20243d}. By jointly leveraging complementary information from different modalities, multimodal models can often outperform unimodal counterparts.

Among various multimodal settings, image–tabular learning has recently gained increasing attention due to its broad applicability in domains such as clinical decision support \cite{borsos2024predicting,bai2020population} and marketing research \cite{huang2022dvm}. 
Images typically capture rich visual patterns, while tabular data provide complementary structured information, such as demographic attributes, numerical measurements, and contextual records.
However, modality incompleteness is common in practice. For example, in healthcare, imaging exams may be unavailable for some patients due to cost or accessibility constraints, while clinical measurements can be missing under emergency conditions. Most existing image–tabular methods \cite{du2024tip,wolf2022daft,hager2023best,duenias2025hyperfusion,huang2024predicting} typically assume that all modalities are available during both training and inference, making them vulnerable in missing-modality scenarios.  Therefore, developing robust image–tabular classification models that can effectively handle missing modalities is of great practical importance.

Incomplete image–tabular learning is non-trivial because the two modalities are not only complementary but also highly heterogeneous, making robust fusion under incompleteness especially difficult. To address this issue, prior image–tabular approaches often aim to learn modality-invariant representations. For instance, DrFuse \cite{yao2024drfuse} performs shared–specific disentanglement and coarse-grained alignment on globally averaged token representations. Similar strategies have been explored in other multimodal domains, such as semantic segmentation \cite{wang2023multi}, emotion recognition \cite{zuo2023exploiting}, and disease diagnosis \cite{chen2023disentangle}. However, these approaches ignore token-level semantic inconsistencies and fine-grained distribution discrepancies, resulting in potential representation entanglement and suboptimal cross-modal alignment. To facilitate cross-modal learning, a recent work \cite{qian2025decalign} introduces a prototype-based distribution alignment, but its prototype modeling is primarily confined to modality-specific representations, without explicitly modeling shared prototypes or class-level semantic consistency. Another line of research addresses missing modalities via knowledge distillation \cite{wang2023learnable,li2024correlation,zhao2024maskmentor}, where a teacher model trained on complete modalities supervises a student model operating on partial modalities. However, this paradigm requires access to all modalities for teacher training, which is impractical when large-scale, complete data are unavailable. Recently, prompt-based methods \cite{shi2024deep,lee2023multimodal} have leveraged prompt-tuning techniques to effectively transfer the capabilities of large vision–language models (\eg, CLIP \cite{radford2021learning}) pre-trained on complete multimodal datasets to tasks involving missing modalities. Despite their success in vision–language tasks, adapting these models to image–tabular settings is non-trivial due to the intrinsic heterogeneity of tabular data and the lack of universally pre-trained image–tabular models. 

To address these challenges, we propose DFPL, a novel framework for incomplete tabular–image classification that disentangles multimodal features into shared and modality-specific prototypes, enabling fine-grained cross-modal interactions at the prototype level while enhancing robustness to incomplete inputs, as illustrated in Fig. \ref{fig:challenge}(b). In detail, we propose a SSPM module to extract compact and diverse shared and modality-specific prototypes via different prototype queries. These prototypes serve as structured semantic summaries of token-level features, which are then used for fine-grained disentanglement and alignment. In addition, we develop a PFA module that enforces prototype-level distribution matching for both shared and modality-specific prototypes, as well as prototype-to-class semantic alignment. Specifically, the former matches cross-modal shared/specific prototypes based on optimal transport (OT), while the latter aligns cross-modal similarity distributions between shared prototypes and class prototypes, thus promoting fine-grained distributional and semantic consistency across modalities. Inspired by the advantages of label-query mechanisms \cite{liu2021query2label} in extracting class-related features, we further introduce a CMA module that adaptively integrates shared and modality-specific representations across two scales (\ie,  global and prototype) by learnable class queries to produce robust classification results. The main contributions of this paper are summarized as follows:

\begin{itemize}
    \item We introduce DFPL, a novel framework for incomplete tabular-image classification that achieves disentangled representation learning from coarse global features to fine-grained prototypes. Extensive experiments on three image-tabular datasets demonstrate its superior performance under various missing-modality scenarios. 
    \item We propose a SSPM module to extract compact and diverse shared and modality-specific prototypes from token-level features, enabling prototype-level disentanglement.
    \item We develop a PFA module that performs cross-modal distribution matching over shared and modality-specific prototypes, and further enforces similarity-distribution consistency between shared prototypes and class prototypes.
    \item We design a CMA module to integrate global- and prototype-level representations in a class-guided manner, improving robustness to missing modalities.
\end{itemize}

\section{Related Work}

\noindent\textbf{Multimodal Image–Tabular Learning.}
Multimodal image–tabular learning has attracted increasing attention in a variety of domains, particularly in the medical field~\cite{polsterl2021combining,xue2024ai,zhou2025glcp, fu2025unleashing}, where images capture visual patterns while tabular data encode structured attributes such as demographics, laboratory tests, and clinical history.
Early works mainly focused on designing feature fusion strategies \cite{wolf2022daft}, such as feature concatenation or attention-based weighting, without explicitly addressing the missing-modality problem.
More recently, self-supervised pre-training \cite{huang2024predicting,delgrange2024a} has been explored to improve cross-modal representation learning.
For example, MMCL~\cite{hager2023best} was the first to leverage contrastive pre-training on large-scale image–tabular pairs, followed by supervised fine-tuning on labeled datasets.
TIP \cite{du2024tip} further enhanced the representation capability by modeling missing variables within tabular data during pre-training.  The emergence of tabular foundation models has recently opened a new direction for image–tabular learning. Compared with conventional tabular encoders, which are usually optimized from scratch on task-specific datasets, tabular foundation models aim to provide transferable representations by exploiting large-scale pretraining or meta-learning over diverse tabular tasks. TIME \cite{luo2025time} leveraged TabPFN \cite{hollmann2022tabpfn}  as a frozen tabular encoder to generate robust and informative tabular embeddings, and then combined them with image features from pretrained vision backbones. MultiModalPFN \cite{kim2026multimodalpfn} further generalized TabPFN from tabular-only settings to multimodal inputs, including tabular data paired with images or text. By mapping non-tabular embeddings into balanced token representations, it mitigates the attention imbalance problem and improves multimodal fusion across both medical and general-purpose benchmarks.

Although these methods significantly improve downstream task performance by exploiting image–tabular correlations, they still assume that both modalities are always available during inference, making them less effective in real-world scenarios where modality missingness is common.

\begin{figure*}[t]
  \centering
   \includegraphics[width=1\linewidth]{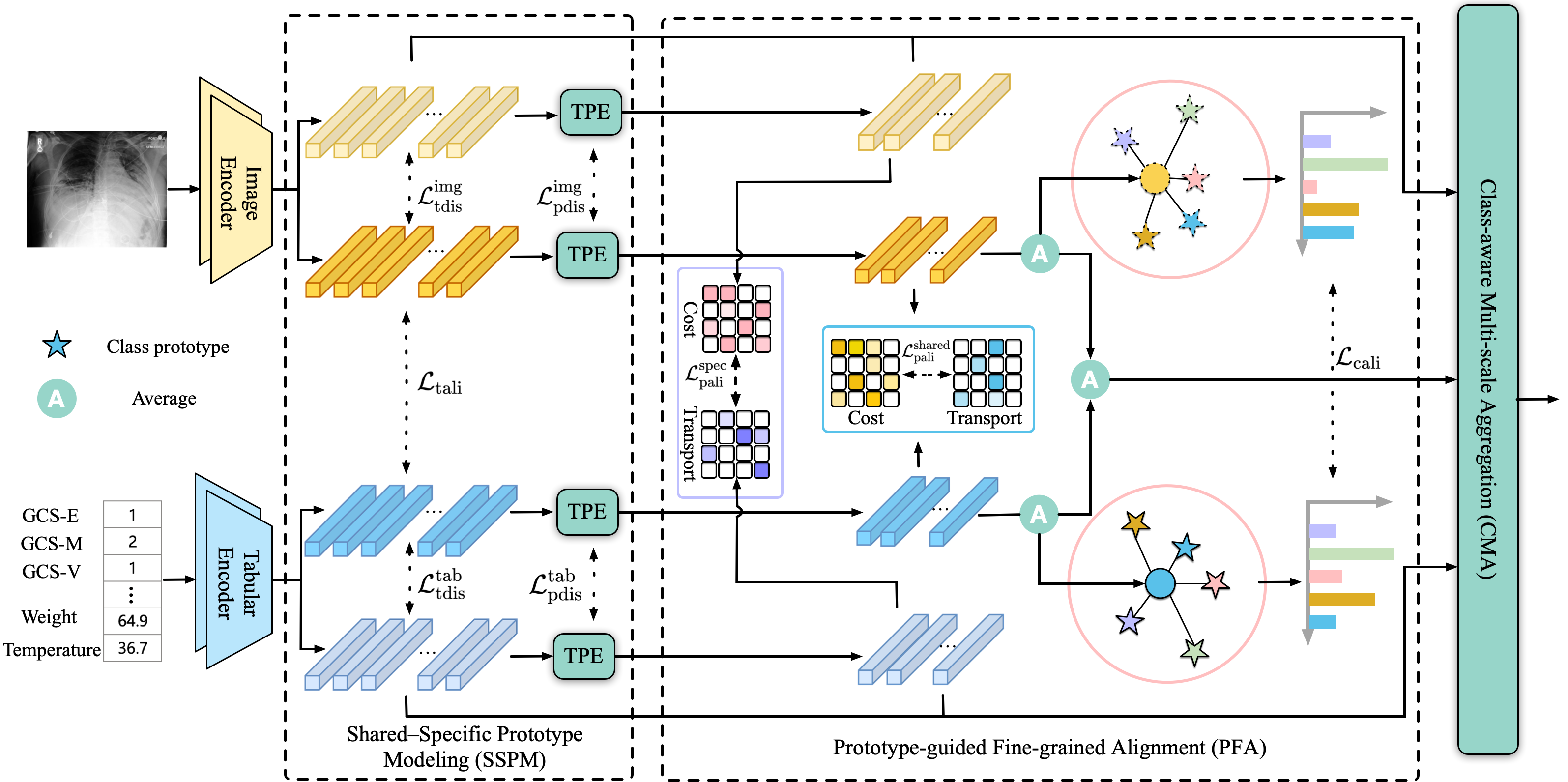}
   \caption{The overview of DFPL. Given multimodal inputs, image and tabular encoders extract modality-specific and shared token-level representations. SSPM aims to extract shared and modality-specific prototypes by TPE for prototype-level disentanglement ($\mathcal{L}_{\text{pdis}}^{m}$, $m \in \{ \text{img, tab}\}$). PFA then performs prototype-level distribution matching ($\mathcal{L}_{\text{pali}}^{\text{shared}}$, $\mathcal{L}_{\text{pali}}^{\text{spec}}$) for both shared and modality-specific prototypes, as well as prototype-to-class semantic alignment ($\mathcal{L}_{\text{cali}}$).
   CMA then adaptively fuses shared and specific information from two scales for class-aware robust prediction. $\mathcal{L}_{\text{tdis}}^{m}$ and $\mathcal{L}_{\text{tali}}$ denote the global-level disentanglement and alignment losses, respectively.
}
\label{fig:framework}
\end{figure*}

\noindent\textbf{Missing Modality in Multimodal Learning.}
The missing-modality problem is prevalent in real-world multimodal scenarios due to acquisition cost, sensor failure, privacy constraints, or incomplete data collection \cite{karimijafarbigloo2024mmcformer,li2025rosa}. Existing solutions can be broadly grouped into modality reconstruction, knowledge distillation, prompt-based adaptation, and modality-invariant representation learning.

Modality reconstruction methods \cite{zhang2024unified,zeng2024missing} attempt to synthesize or impute the missing modality from the available ones using autoencoders, generative adversarial networks, diffusion models, or cross-modal generation modules. Although these methods provide an intuitive way to recover missing information, they may introduce noisy or biased synthetic features when the cross-modal mapping is ambiguous. This issue becomes more severe in image-tabular learning, where visual patterns and structured variables are highly heterogeneous and do not necessarily have one-to-one correspondences.

Knowledge distillation-based approaches \cite{wang2023learnable,li2024correlation,zhao2024maskmentor} train a teacher network with complete modalities to guide a student model that operates on partial modalities. By transferring the predictions or intermediate representations of the complete-modality teacher, these methods can improve the performance of unimodal or incomplete-modality students. However, they usually require fully paired multimodal data to train a reliable teacher model, which is often unavailable in practical applications.

Prompt-based methods \cite{10888848,shi2024deep,lee2023multimodal} leverage large vision-language models by designing task-specific prompts to handle missing inputs. These methods have shown promising results in vision-language tasks, where large-scale pre-trained models provide strong cross-modal priors. Nevertheless, they are difficult to directly adapt to image-tabular learning, because tabular data are schema-specific, domain-dependent, and structurally different from natural language. Moreover, general-purpose pre-trained image-tabular models are still lacking, limiting the transferability of prompt-based solutions to heterogeneous image-tabular scenarios.

Modality-invariant representation learning seeks to project available modalities into a shared latent space, where the learned representations preserve useful cross-modal cues and enable inference when one modality is missing \cite{yao2024drfuse,wang2023multi,zuo2023exploiting,chen2023disentangle}. Recent methods further introduced shared-specific disentanglement to separate modality-shared information from modality-specific characteristics. Despite their effectiveness, most of them perform alignment on globally aggregated features, which mainly capture coarse-grained cross-modal consistency. As a result, token-level semantic inconsistencies and fine-grained distribution discrepancies between heterogeneous modalities may remain unresolved, leading to representation entanglement and suboptimal cross-modal alignment.

Different from these methods, the proposed DFPL focuses on fine-grained shared-specific prototype learning for incomplete image--tabular classification. Instead of aligning only global representations or reconstructing missing modalities, DFPL explicitly extracts shared and modality-specific prototypes from token-level features, performs prototype-level distribution matching, and further enforces class-level semantic consistency. This design enables the model to better exploit distributional and semantic cues from the available modality, thereby improving robustness under various missing-modality scenarios.

\section{Method}
\label{sec:method}

\subsection{Problem Formulation and Overall Framework}\label{sec:problem}
\noindent\textbf{Problem Formulation.}  
We consider a multimodal classification setting with  image $\mathbf{x}^{\text{img}} \in \mathbb{R}^{H \times W \times 3}$ and tabular $\mathbf{x}^{\text{tab}}$, where $H$ and $W$ denote the image height and width, respectively.  
The tabular modality may appear as a static feature vector $\mathbf{x}^{\text{tab}} \in \mathbb{R}^{L}$ or a temporal sequence $\mathbf{x}^{\text{tab}} \in \mathbb{R}^{S \times L}$ (e.g., electronic health records (EHR)), where $S$ is the temporal length and  $L$ is the feature dimension.  
Given a dataset $\mathcal{D} = \{(\mathbf{x}^{\text{img}}_n, \mathbf{x}^{\text{tab}}_n, \mathbf{y}_n)\}_{n=1}^N$ 
with $N$ samples, the label vector $\mathbf{y}_n \in \{0,1\}^C$ 
denotes the class annotations, where $C$ is the number of classes. Missing modalities are indicated by a binary mask $\mathbf{m} = [m_I, m_T] \in \{0,1\}^2$,  where $m_I=1$ if the image is present.  Similar to \cite{yao2024drfuse,wang2023multi}, missing inputs are replaced with zero vectors during both training and inference.  

\noindent\textbf{Overall Framework.} Following the recent work~\cite{yao2024drfuse} on missing-modality learning, we employ, for each modality, two encoders with identical architectures to extract shared and specific token-level features, \ie, $\mathbf{T}^{m}_{\text{shared}}, \mathbf{T}^{m}_{\text{spec}} \in \mathbb{R}^{L_{m} \times D}$, where modality $m \in \{ \text{img, tab}\}$,  $D$ is the hidden dimension and $L_{m}$ represents the token lengths of $m$. The two encoders share lower-layer parameters to capture common low-level patterns, while maintaining separate higher layers to specialize in shared or modality-specific semantics. As shown in Fig.~\ref{fig:framework},  SSPM jointly learns compact and diverse shared and specific prototypes, followed by fine-grained disentanglement (Sec. \ref{sec:SSPM}).  
Besides, PFA aims to further align prototype-level and prototype-to-class distributions, achieving distributional and semantic consistency across modalities (Sec. \ref{sec:PFA}).  
Finally, CMA is introduced to fuse shared and specific information to yield robust predictions (Sec. \ref{sec:CMA}).


\subsection{Shared-Specific Prototype Modeling (SSPM)}\label{sec:SSPM}
As aforementioned, prior shared–specific models \cite{yao2024drfuse,du2025stil} operate on globally averaged features, capturing coarse modality-invariant cues but overlooking fine-grained token semantics. A straightforward solution would be to enforce disentanglement constraints on all token-level features directly, but the large number of tokens may introduce noisy and redundant supervision. In contrast, our SSPM learns a small set of semantic prototypes to organize token features into compact shared and modality-specific components.


\noindent\textbf{Coarse-grained Shared-Specific Modeling.}
We first obtain global representations by averaging token features within each modality branch. The objective is to disentangle the shared and specific subspaces of each modality while aligning shared representations across modalities at a coarse level. Following \cite{yao2024drfuse}, the loss is defined as:
\begin{equation}
\label{eq:cdis} 
\begin{aligned}
\mathcal{L}_{\text{cssm}} = 
\sum_{m\in \{\text{img},\text{tab}\}}
\frac{\left|\mathbf{t}_{\text{shared}}^{m}\cdot(\mathbf{t}_{\text{spec}}^{m})^\mathrm{T}\right|}
{||\mathbf{t}_{\text{shared}}^{m}||_2 \, ||\mathbf{t}_{\text{spec}}^{m}||_2}
+\text{JSD}\!\left(\mathbf{t}_{\text{shared}}^{\text{img}}, \mathbf{t}_{\text{shared}}^{\text{tab}}\right),
\end{aligned}
\end{equation}
\noindent where the first and second items are global disentanglement (\ie, $\mathcal{L}_{\text{tdis}}$) and alignment (\ie, $\mathcal{L}_{\text{tali}}$) losses, respectively. $\mathbf{t} \in \mathbb{R}^{D}$ is the global average of $\mathbf{T}$, and $\text{JSD}(\cdot)$ is the Jensen–Shannon divergence. This coarse-grained modeling defines global semantic boundaries that can effectively guide and stabilize prototype-level refinement (verified in \textbf{Supp. C}).

\noindent\textbf{Token Prototype Extractor (TPE).} Inspired by prototype learning~\cite{snell2017prototypical,luo2025exploring} that represents each semantic concept or class with a compact set of learnable prototypes, we propose to dynamically extract a small number of semantic prototypes from the token-level features of each modality branch through a learnable query-based attention mechanism, where each prototype compactly summarizes a distinct latent semantic pattern. Formally, given token-level representations 
$\mathbf{T}\in \mathcal{T}=\left\{ \mathbf{T}^{m}_{\text{shared}}, \mathbf{T}^{m}_{\text{spec}} \mid m \in \left\{\text{img}, \text{tab}\right\} \right\}$, 
we introduce $K$ learnable prototype queries for each branch, denoted as 
$\mathbf{F}\in \mathcal{F}=\left\{ \mathbf{F}^{m}_{\text{shared}}, \mathbf{F}^{m}_{\text{spec}} \mid m \in \left\{\text{img}, \text{tab}\right\} \right\}$, 
to produce shared and modality-specific prototypes 
$\mathbf{P}\in \mathcal{P}=\left\{ \mathbf{P}^{m}_{\text{shared}}, \mathbf{P}^{m}_{\text{spec}} \mid m \in \left\{\text{img}, \text{tab}\right\} \right\}$. 
The four groups of prototype queries in $\mathcal{F}$ are parameter-independent, which encourages different branches to focus on distinct semantic patterns and facilitates subsequent shared--specific disentanglement. 
Mathematically, each query and its corresponding token-level features are fed into a branch-specific attention module to extract prototypes:
\begin{equation}
\label{eq:tpe} 
\begin{aligned}
\mathbf{F^{\prime } }&= \text{softmax}\left(\frac{\mathbf{Q}  \mathbf{K}^\top}{\sqrt{D}} \right)\mathbf{V}+\mathbf{F},\\
\mathbf{P}&=FFN(\mathbf{F^{\prime } })+\mathbf{F^{\prime } },
\end{aligned}
\end{equation}
where $\mathbf{Q}=\mathbf{FW}_{q}$, $\mathbf{K}=\mathbf{TW}_{k}$, and $\mathbf{V}=\mathbf{TW}_{v}$. $\mathbf{W}_{q}$. $\mathbf{W}_{k}$, $\mathbf{W}_{v}\in \mathbb{R}^{D \times D} $ are learnable projection matrices. $FFN(\cdot)$ represents the feed-forward network.

\noindent\textbf{Prototype Diversity Loss.} After obtaining the prototype-level representation  $\mathbf{P}\in \mathbb{R}^{K\times D} 
$, it is crucial to ensure that each prototype within it encodes distinct semantics to avoid redundant prototypes. To this end, we introduce a diversity regularization loss that minimizes cosine similarity (\ie, $\cos(\cdot)$) between every pair of prototypes. Given $\mathbf{P}$, we have:
\begin{equation}
\label{eq:div} 
\mathcal{L}_{\text{div}} = \frac{1}{K (K - 1)} \sum_{\mathbf{P} \in \mathcal{P}} \sum_{i \neq j} \max \big(0, \cos(\mathbf{P}_{i}, \mathbf{P}_{j}) - \delta \big),
\end{equation}
where $\delta \in [0,1]$ is a similarity margin, which enforces separation only when two prototypes are overly close. This constraint encourages each prototype to attend to unique latent semantics and avoids mode collapse.

\noindent\textbf{Prototype-level Disentanglement.} We further enforce disentanglement between shared and modality-specific representations at the prototype level after extracting the shared-specific prototype set $\mathcal{P}$. Since $K$ is relatively small compared with the original number of tokens, we directly average over prototype-level representations, \ie, $\mathbf{P}_{\text{shared}}^{m}$ and $\mathbf{P}_{\text{spec}}^{m}$  to obtain $\mathbf{p}_{\text{shared}}^{m}$ and $\mathbf{p}_{\text{spec}}^{m}$ respectively, where $m \in\left \{\text{img, tab}  \right \}$. We then apply the same loss defined in Eq. (\ref{eq:cdis}) to $\mathbf{p}_{\text{shared}}^{m}$ and $\mathbf{p}_{\text{spec}}^{m}$ to get $\mathcal{L}_{\text{pdis}}$. By combining coarse-grained and prototype-level disentanglement losses, we achieve a clearer separation between shared and modality-specific spaces.


\subsection{Prototype-guided Fine-grained Alignment (PFA)}\label{sec:PFA}
Despite global-level alignment in Eq. (\ref{eq:cdis}), cross-modal distribution and semantic discrepancies remain due to high modality heterogeneity. Inspired by the success of fine-grained alignment in vision-language learning \cite{xiefg,shuilarge}, which shows that aligning specific regions within images with corresponding text can enhance representation learning, we propose to boost fine-grained distributional and
semantic consistency across modalities by aligning prototype-level distributions, as well as prototype-to-class semantics.

\noindent\textbf{Prototype-level Distribution Alignment.} Unlike the region–text pairs in ~\cite{xiefg,shuilarge}, the shared and specific prototypes are not in a one-to-one correspondence across modalities, making direct prototype-level alignment infeasible. Therefore, we formulate the alignment as an OT problem, where the OT distance~\cite{cuturi2013sinkhorn,chizat2020faster,nguyen2022improving} is employed to measure the distributional discrepancy between prototypes across modalities. Given modality-shared prototypes 
$\mathbf{P}^{\text{img}}_{\text{shared}}$ and 
$\mathbf{P}^{\text{tab}}_{\text{shared}}$, 
we construct a cost matrix that measures the distance between each prototype pair:
\begin{equation}
\mathbf{C}_{ij} = \| \mathbf{P}^{\text{img}}_{\text{shared},i} - \mathbf{P}^{\text{tab}}_{\text{shared},j} \|_2^2.
\label{eq:ot_cost}
\end{equation}

We seek an optimal transport plan $\mathbf{G}\in\mathbb{R}^{K\times K}$ that minimizes the transport cost across modalities. The entropic-regularized OT solution is efficiently approximated using the Sinkhorn algorithm \cite{cuturi2013sinkhorn}, and the resulting prototype-level distribution alignment loss is formulated as:
\begin{equation}
\mathcal{L}_{\text{pali}}^{\text{shared}}
= \min_{\mathbf{G} \in \Pi(\mu_{\text{img}},\mu_{\text{tab}})} \langle \mathbf{G}, \mathbf{C} \rangle 
+ \varepsilon \sum_{i,j} \mathbf{G}_{ij}\log \mathbf{G}_{ij},
\label{eq:sinkhorn}
\end{equation}
where  $\Pi(\mu_{\text{img}},\mu_{\text{tab}})=\{\mathbf{G} \,|\, \mathbf{G}\mathbf{1}_K=\mu_{\text{img}},\, \mathbf{G}^\top \mathbf{1}_K=\mu_{\text{tab}}\}$ denote the marginal constraints of total mass equality between marginal distributions, \ie, $\mu_{\text{img}}$ and $\mu_{\text{tab}}$. $\mathbf{1}_K$ represents a $K$-dimensional vector of ones. $\langle \mathbf{G},\mathbf{C}\rangle=\sum_{i,j}\mathbf{G}_{ij}\mathbf{C}_{ij}$ is the total transportation cost.  Weight $\varepsilon$ controls the regularization strength, which is set to 0.1.

Although modality-specific prototypes encode modality-dependent characteristics, the structural gap between image and tabular subspaces can still be substantial due to their heterogeneous feature distributions. If left unregularized, this discrepancy may hinder effective cross-modal interaction and weaken robustness under missing-modality scenarios. To this end, we further extend the Sinkhorn-based OT to modality-specific prototypes, yielding $\mathcal{L}_{\text{pali}}^{\text{spec}}$. Moderate alignment encourages structural compatibility across modality-dependent subspaces while preserving their complementary characteristics. The overall distribution alignment loss is then defined as $\mathcal{L}_{\text{pali}}=\mathcal{L}_{\text{pali}}^{\text{shared}} +\omega \mathcal{L}_{\text{pali}}^{\text{spec}}
$, where $\omega$ is a small weighting factor to prevent over-alignment that may collapse modality-specific characteristics. 

\noindent\textbf{Prototype-to-Class Semantic Alignment.} 
While prototype-level alignment reduces the discrepancy between modalities in the prototype space, it does not explicitly enforce semantic consistency with respect to the class space. Therefore, we propose to align cross-modal shared semantics by extracting multi-level class prototypes and 
encouraging prototype-to-class semantic consistency. This design enhances the semantic structure and robustness of shared representations.

Class prototypes are the representative embeddings of all positive samples in each class within a mini-batch. Unlike prior works \cite{du2025stil,li2024correlation}, we construct class prototypes at both global and prototype levels and fuse them with a learnable gate to provide richer semantic cues for alignment. For each class $c$ that appears in the current mini-batch, the class prototype $\mathbf{\nu }_{c}^{m}$ for modality $m$ is constructed by fusing the global prototype $\mathbf{\nu }_{\text{glo}, c}^{m}$ with the prototype-level counterpart $\mathbf{\nu }_{\text{proto}, c}^{m}$:
\begin{equation}
\label{eq:proto_class}
\begin{aligned}
\mathbf{\nu }_{c}^{m} &= \lambda_c \mathbf{\nu }_{\text{glo}, c}^{m}+ (1-\lambda_c)\mathbf{\nu }_{\text{proto}, c}^{m} \\
&=\lambda_c \frac{1}{|\mathcal{B}_{c}|} \sum_{i \in \mathcal{B}_{c}} (\mathbf{t}^{m}_{\text{shared},i})+ (1-\lambda_c) \frac{1}{|\mathcal{B}_{c}|} \sum_{i \in \mathcal{B}_{c}} (\mathbf{p}^{m}_{\text{shared},i}),
\end{aligned}
\end{equation}
where $\mathcal{B}_c = \{ i \mid \mathbf{y}_{i,c} = 1 \}$  the index set of positive samples in the mini-batch, and $\lambda_c$ balances global- and prototype-level cues. Rather than using a fixed $\lambda_c$, a gating mechanism adaptively learns the relative importance of two prototype types, which is defined as:
\begin{equation}
\label{eq:proto_weight}
\begin{aligned}
\lambda_{c} = \sigma \left( \mathbf{w}^{\top} \left[ \mathbf{\nu }_{\text{glo},c}^{m} \, ; \, \mathbf{\nu }_{\text{poto},c}^{m} \right] \right),
\end{aligned}
\end{equation}
where $\sigma(\cdot)$ is the sigmoid function, 
$\mathbf{w} \in \mathbb{R}^{2D}$ is a learnable gating vector, 
and $[\cdot ; \cdot]$ denotes concatenation. For each modality, we then derive the class-aware similarity distribution by calculating the cosine similarity between the aggregated shared prototype (\eg, $\mathbf{p}_{\text{shared}}^{\text{img}}$) and all class prototypes. Then, prototype-to-class semantic consistency is defined as:
\begin{equation}
\label{eq:proto-to-class}
\mathcal{L}_{\text{cali}} = \frac{1}{|\mathcal{C}|} \sum_{c \in \mathcal{C}}\left\| \cos(\mathbf{p}^{\text{img}}_{\text{shared}},\mathbf{\nu }_{c}^{\text{img}}) - \cos(\mathbf{p}^{\text{tab}}_{\text{shared}},\mathbf{\nu }_{c}^{\text{tab}}) \right\|_2^2,
\end{equation}
where $\mathcal{C}$ denotes the set of all classes present in the mini-batch. The overall PFA loss $\mathcal{L}_{\text{pfa}}$ is obtained as the sum of $\mathcal{L}_{\text{pali}}$ and $\mathcal{L}_{\text{cali}}$, jointly enforcing fine-grained distributional and semantic consistency in shared–specific and class prototype spaces, respectively, to improve cross-modal reasoning under missing modalities.


\subsection{Class-aware Multi-scale Aggregation (CMA)}\label{sec:CMA}
In heterogeneous image-tabular learning, modality importance is often sample- and class-dependent. For example, in clinical diagnosis, different diseases may rely on different combinations of imaging evidence and structured patient attributes \cite{yao2024drfuse}. Existing shared-specific modeling methods typically fuse disentangled representations through concatenation \cite{wang2023multi} or cross-attention \cite{du2025stil}, but they usually overlook class-dependent modality preferences. Different from \cite{yao2024drfuse}, which builds queries and class-wise keys from disentangled features, we introduce learnable class queries that directly attend to different representation types (shared/specific) and scales (global/prototype). In this way, CMA adaptively integrates global- and prototype-level shared/specific cues according to class-related requirements, thereby improving robustness under missing-modality scenarios.

Specifically, as the extracted shared prototypes encode modality-invariant semantic information, we construct a fused shared prototype to aggregate the available modality cues. Given the modality availability indicator $\mathbf{m}$, the fused shared prototype is defined as:
\begin{equation}
\begin{aligned}
\mathbf{p}_{\text{shared}}^{\text{tab-img}}=\left\{\begin{matrix}
  (\mathbf{p}_{\text{shared}}^{\text{tab}}+\mathbf{p}_{\text{shared}}^{\text{img}})/2&, \text{if} \  \mathbf{m}=[1,1] \\
  \mathbf{p}_{\text{shared}}^{\text{tab}}& , \text{if} \  \mathbf{m}=[0,1] \\
  \mathbf{p}_{\text{shared}}^{\text{img}}& , \text{if} \  \mathbf{m}=[1,0] ,
\end{matrix}\right.
\end{aligned}
\label{eq:shared_ave}
\end{equation}
where the shared prototypes are averaged when both modalities are available; otherwise, the prototype of the available modality is directly adopted. Similarly, we can get the fused global feature $\mathbf{t}_{\text{shared}}^{\text{tab-img}}$ by Eq. (\ref{eq:shared_ave}). We then construct a complete multi-scale representation $\mathbf{H} \in \mathbb{R}^{6 \times D}$ as:
\begin{equation}
\begin{aligned}
\mathbf{H} = \left[\mathbf{t}_{\text{spec}}^{\text{img}};\mathbf{p}_{\text{spec}}^{\text{img}}; \mathbf{t}_{\text{shared}}^{\text{tab-img}}; \mathbf{p}_{\text{shared}}^{\text{tab-img}};   \mathbf{t}_{\text{spec}}^{\text{tab}};\mathbf{p}_{\text{spec}}^{\text{tab}} \right].
\end{aligned}
\label{eq:proto_concat}
\end{equation}

Given the multi-scale representation $\mathbf{H}$ and the learnable class queries $\mathbf{Q}_{\text{class}}\in\mathbb{R}^{C\times D}$, we perform masked cross-attention to obtain class-aware features. Specifically, $\mathbf{Q}_{\text{class}}$ serves as the query, while $\mathbf{H}$ serves as both the key and value. The attention process is formulated as: 
\begin{equation}
\begin{aligned}
\mathbf{Q}=\mathbf{Q}_{\text{class}}\mathbf{W}_{Q},\quad
\mathbf{K}=\mathbf{H}\mathbf{W}_{K},\quad
\mathbf{V}=\mathbf{H}\mathbf{W}_{V}, \\
\mathbf{Z}
=
\text{softmax}
\left(
\frac{\mathbf{Q}\mathbf{K}^{\top}}{\sqrt{D}}
+
\mathbf{M}
\right) \mathbf{V},
\label{eq:cross_attn}
\end{aligned}
\end{equation}
where $\mathbf{W}_{Q}$, $\mathbf{W}_{K}$, and $\mathbf{W}_{V} $ are learnable projection matrices. The modality-aware mask $\mathbf{M}\in\mathbb{R}^{C\times 6}$ is broadcast from the modality availability indicator, where entries corresponding to missing-modality representations are set to -$\infty$ and others are set to $0$. In this way, the softmax operation assigns zero attention weights to unavailable representations, preventing them from contributing to the class-aware aggregation. The class-aware features $\mathbf{Z}$ are then passed to a linear classifier for the final prediction $\hat{\mathbf{y}}$ using binary cross-entropy 
\begin{equation}
\mathcal{L}_{\text{cls}}
=
-\frac{1}{C}
\sum_{c=1}^{C}
\left[
\mathbf{y}_{c}\log \hat{\mathbf{y}}_{c}
+
(1-\mathbf{y}_{c})\log (1-\hat{\mathbf{y}}_{c})
\right].
\label{eq:cls_loss}
\end{equation}

Finally, the overall training objective to be minimized is defined as follows:
\begin{equation}
\mathcal{L}_{\text{total}} = \mathcal{L}_{\text{cls}} + \alpha \mathcal{L}_{\text{cssm}} + \beta (\mathcal{L}_{\text{div}} + \mathcal{L}_{\text{pdis}}) + \gamma \mathcal{L}_{\text{pfa}},
\label{eq:total_loss}
\end{equation}
where $\alpha, \beta, \gamma$ are hyper-parameters balancing the contribution of each component. Following ~\cite{wang2023multi}, the losses associated with the missing modality are omitted during training. More training and inference details are provided in Supp. \textbf{A}.

\begin{table*}[!t]
  \renewcommand{\arraystretch}{0.8} 
  \centering
  \caption{Comparison results on the MIMIC, ADNI, and DVM datasets under various missing-modality cases with different missing rates applied to both training and testing phases. The bold number indicates the best performance. }
  \label{tab:comapare_sota}
  \begin{tabular}{c|c|cc|cccccccc}
      \toprule 
      \multirow{2}{*}{ Datasets } & \multirow{2}{*}{\makecell[c]{  Missing  rate $\eta$}} & \multicolumn{2}{c|}{ Train/Test}  & \multicolumn{8}{c}{ Method}\\
       & & Image & Tabular  & TIP  & IF-MMIN  & STiL & IM-Fuse & DMRNet & ShaSpec   & DrFuse  & Ours  \\
      \midrule 
      \multirow{9}{*}{\makecell[c]{ 
       MIMIC \\(PRAUC)}} & \multirow{3}{*}{$30 \%$} & $100 \%$ & $70 \%$  &37.72 & 38.45& 38.95& 37.88 & 38.20 & 38.94 & 39.30 & \textbf{40.63} \\
      & & $70 \%$ & $100 \%$ & 40.10	&40.76	&41.00	&40.61 & 40.77	&41.21	&41.60	&\textbf{42.73}\\
      & & $85 \%$ & $85 \%$ & 39.23	&40.04	&40.12	&39.39 & 39.91	&40.14	&40.34	&\textbf{41.67} \\
    \cmidrule{2-12}
      & \multirow{3}{*}{$50 \%$} &  $100 \%$ & $50 \%$ & 34.56	& 35.09	& 35.70	& 35.11 & 35.10	& 35.84	& 36.42	& \textbf{38.43}  \\
      & & $50 \%$ & $100 \%$ & 39.93	& 40.34	& 40.95	&40.30 & 40.33	&40.96	&41.22	&\textbf{42.25} \\
       & & $75 \%$ & $75 \%$ & 36.34	&37.10	&37.45	&36.98	& 36.90  &37.13	&38.64	&\textbf{40.22}  \\
      \cmidrule {2-12}
      & \multirow{3}{*}{$70 \%$} & $100 \%$ & $30 \%$ & 32.00	 &33.07 &33.80	 &32.34 & 33.51 	 &33.89	 &34.50	 &\textbf{36.00} \\
      & & $30 \%$ & $100 \%$ & 39.02	 &40.00	 &40.54 &39.48 & 39.64	 &40.67	 &41.11	 &\textbf{42.06}\\
      & & $65 \%$ & $65 \%$ & 34.01	 &34.56	 &35.28	 &34.23 & 34.56	 &35.64	 &36.20	 &\textbf{37.62}  \\
      \midrule 
      \multirow{9}{*}{\makecell[c]{ADNI \\(ACC)}} & \multirow{3}{*}{$30 \%$} & $100 \%$ & $70 \%$ &72.59	&73.53	&74.30	&72.12	&73.21	&74.14	&74.61 &\textbf{75.86}\\
      & & $70 \%$ & $100 \%$ &86.45	&86.60	&86.92	&85.20	&86.14	&86.76	&86.60	&\textbf{87.07}\\
      & & $85 \%$ & $85 \%$ &77.57	&78.35	&78.82	&77.10	&78.19	&79.13	&79.91	&\textbf{80.84}\\
      \cmidrule{2-12}
      & \multirow{3}{*}{$50 \%$} & $100 \%$ & $50 \%$&62.31	&63.40	&64.02	&61.68	&62.93	&63.71	&64.49	&\textbf{65.89} \\
      & & $50 \%$ & $100 \%$ &86.14	&85.83	&86.29	&85.20	&85.36	&86.14	&86.29	&\textbf{86.92} \\
      & & $75 \%$ & $75 \%$ &72.90	&73.52	&74.61	&72.12	&73.05	&74.30	&74.61	&\textbf{76.48}  \\
    \cmidrule{2-12}
      & \multirow{3}{*}{$70 \%$} & $100 \%$ & $30 \%$ &55.14	&55.30	&55.76	&54.83	&55.45	&56.54	&57.17	&\textbf{59.03}\\
      & & $30 \%$ & $100 \%$ &85.83	&85.51	&85.98	&84.89	&85.05	&86.14	&86.14	&\textbf{86.76}\\
      & & $65 \%$ & $65 \%$ &70.09	&71.18	&71.96	&69.47	&70.56	&71.81	&72.59	&\textbf{74.45} \\
        \midrule 
      \multirow{9}{*}{\makecell[c]{DVM \\(ACC)}} & \multirow{3}{*}{$30 \%$} & $100 \%$ & $70 \%$ &87.64	&88.45	&89.01	&87.98 & 88.87	&89.55	&90.52	&\textbf{95.54}  \\
      & & $70 \%$ & $100 \%$ & 95.11	&95.80	&95.88	&95.60 & 95.65	&96.04	&96.48	&\textbf{98.23}\\
      & & $85 \%$ & $85 \%$ & 89.78	&90.21	&91.45	&90.31 & 90.89	&91.90	&92.91	&\textbf{96.24}  \\
      \cmidrule{2-12}
      & \multirow{3}{*}{$50 \%$} & $100 \%$ & $50 \%$ &86.01	&86.26	&86.89	&85.73 & 86.45	&87.76	&89.03	&\textbf{93.43} \\
      & & $50 \%$ & $100 \%$ & 94.89	&95.44	&95.50	&94.90 & 95.45 	&95.52	&96.00	&\textbf{97.66} \\
      & & $75 \%$ & $75 \%$ &89.00	&89.34	&89.74	&89.12	& 89.40 &90.04	&91.01	&\textbf{94.50}  \\
    \cmidrule{2-12}
      & \multirow{3}{*}{$70 \%$} & $100 \%$ & $30 \%$ & 84.94	&85.14	&85.56 &85.16 & 85.24	&86.46	&87.80	&\textbf{92.59} \\
      & & $30 \%$ & $100 \%$ & 94.00	&94.62 &94.89	&94.26 & 94.40	&95.01	&95.33&\textbf{97.50}\\
      & & $65 \%$ & $65 \%$ &86.86	&87.12	&87.34	&86.96 & 86.91	&87.65 &89.16 &\textbf{92.63}  \\
     \bottomrule
    \end{tabular}
    \end{table*}

\section{Experiment}

\noindent \textbf{Datasets and Evaluation Metrics.} 
We evaluate the proposed DFPL on three different types of image–tabular datasets: MIMIC \cite{johnson2019mimic}, ADNI \cite{jack2008alzheimer}, and DVM \cite{huang2022dvm}.
MIMIC comprises time-series tabular data (EHR) and paired chest radiographs for multi-label disease prediction across 25 diagnostic categories. 
Following the preprocessing in \cite{yao2024drfuse}, we extract paired image–EHR samples within the first 48 hours of ICU admission, where each temporal sequence contains 17 clinical variables. After preprocessing, the dataset is split into 7,637 training, 857 validation, and 2,136 testing samples. Model performance is evaluated using the area under the precision–recall curve (PRAUC). ADNI aims to investigate the diagnosis and progression of Alzheimer’s disease. Following \cite{zhang2024modality,fu2025unleashing}, we mainly collect 1,913 MRI–tabular pairs, where each tabular record contains 43 clinical variables. The subjects span three categories of cognitive status: cognitive normal (CN), mild cognitive impairment (MCI), and dementia (DE). We perform a subject-level random split into 1,531, 191, and 191 subjects for training, validation, and testing, respectively. Since each subject may have multiple longitudinal MRI scans, this split results in 5,880, 783, and 642 scan-level samples for the three sets. The DVM dataset is used for a car model classification task with 283 categories, where each sample contains an RGB image and 17 tabular attributes. The dataset is divided into 70,565 training, 17,642 validation, and 88,207 testing samples. Model performance on ADNI and DVM is evaluated using top-1 classification accuracy (ACC).

\noindent \textbf{Implementation Details.} ResNet-50~\cite{he2016deep} is used as the image encoder, and different transformer-based encoders are adopted for EHR and static tabular data, respectively, following \cite{yao2024drfuse,du2024tip}. The hidden dimension of image and tabular representations is set to 256 for MIMIC and 512 for both ADNI and DVM. The hyper-parameters are set as $\alpha=1$, $\beta=0.1$, $\gamma=0.1$, and $\delta=0.2$. The number of prototypes $K$ is set to 5. All experiments are conducted with batch sizes of 16 (MIMIC and ADNI) and 256 (DVM). For the missing-modality setting, we replace the input with a zero-filled tensor. For a fair comparison, all baseline models follow the same data pre-processing and training protocols. Specifically, the images are resized to $224\times224$ and $128\times128$ for the MIMIC and DVM datasets, respectively. For the ADNI dataset, the 3D MRI scans (\ie, T1) are resized to $128\times128\times128$. All models are trained on two NVIDIA A100 GPUs for 100 epochs using the Adam optimizer. Additional implementation details are provided in Supp. \textbf{B}.


\noindent \textbf{Missing Modality Setting During Training and Testing.} We define the missing rate $\eta\%$ as the proportion of incomplete samples in the dataset. For each dataset, we consider three cases: (i) \textit{tabular missing} and (ii) \textit{image missing}, where $\eta\%$ of the samples contain only one modality and the remaining $1-\eta\%$ contain both; and (iii) \textit{both missing}, where $\tfrac{\eta}{2}\%$ of the samples contain only tabular, $\tfrac{\eta}{2}\%$ contain only images, and the remaining $1-\eta\%$ are complete.

\subsection{Comparison with State-of-the-art Methods} 
We compare our proposed DFPL with seven representative methods, including complete-modality methods: TIP~\cite{du2024tip}, and missing-modality methods: IF-MMIN~\cite{zuo2023exploiting}, ShaSpec~\cite{wang2023multi}, IM-Fuse~\cite{pipoli2025fuse}, STiL~\cite{du2025stil}, DMRNet~\cite{wei2024robust} and DrFuse~\cite{yao2024drfuse}.  Extensive experiments are conducted to thoroughly evaluate the robustness and effectiveness of DFPL under different missing-modality scenarios.  Further implementation details for all baseline methods are provided in Supp. \textbf{B}.

As shown in Table~\ref{tab:comapare_sota}, DFPL achieves the highest PRAUC on MIMIC under all missing-modality settings. It outperforms the complete-modality baseline TIP by 4\% PRAUC when 70\% of tabular inputs are missing, highlighting its effectiveness in handling incomplete modalities. When compared with missing-modality approaches, our method also demonstrates consistent superiority, surpassing the strongest recent competitor, \ie, DrFuse, by 1.33\%, 1.58\%, and 1.42\% under 30\%, 50\%, and 70\% both-modality-missing scenarios, respectively. Notably, the recent IM-Fuse, which employs Mamba~\cite{gu2023mamba} to directly process sparse inputs, performs suboptimally in the highly heterogeneous image–tabular learning setting. These results indicate that fine-grained disentanglement and alignment, together with class-aware multi-scale aggregation, enable our model to effectively capture complementary information even when one modality is largely absent. On the ADNI dataset, DFPL consistently achieves the best accuracy across all missing-modality settings. We further observe a consistent asymmetric robustness pattern on both MIMIC and ADNI: performance degradation is more pronounced when tabular inputs are missing than when images are missing. This phenomenon is expected, as clinical tabular variables often provide more direct and label-relevant cues, while imaging features may contain more heterogeneous variations that are not always tightly coupled with the target labels. Importantly, despite the more challenging tabular-missing scenarios, DFPL maintains clear superiority over competing methods, demonstrating its ability to effectively exploit the remaining modality through disentangled fine-grained prototype learning. 

On the DVM dataset, DFPL significantly outperforms all competing approaches. With 30\% missing tabular data, it reaches 95.54\% accuracy, exceeding DrFuse by 5\%. Moreover, the superiority of our method remains consistent across more challenging scenarios. For instance, with 50\% and 70\% missing rates, our approach still achieves 93.45\% and 92.59\% accuracy, respectively, exceeding the closest baseline by more than 4.0\%. These improvements can be attributed to the enhanced learning of shared and modality-specific representations.

\begin{figure}
  \centering
   \includegraphics[width=1\linewidth]{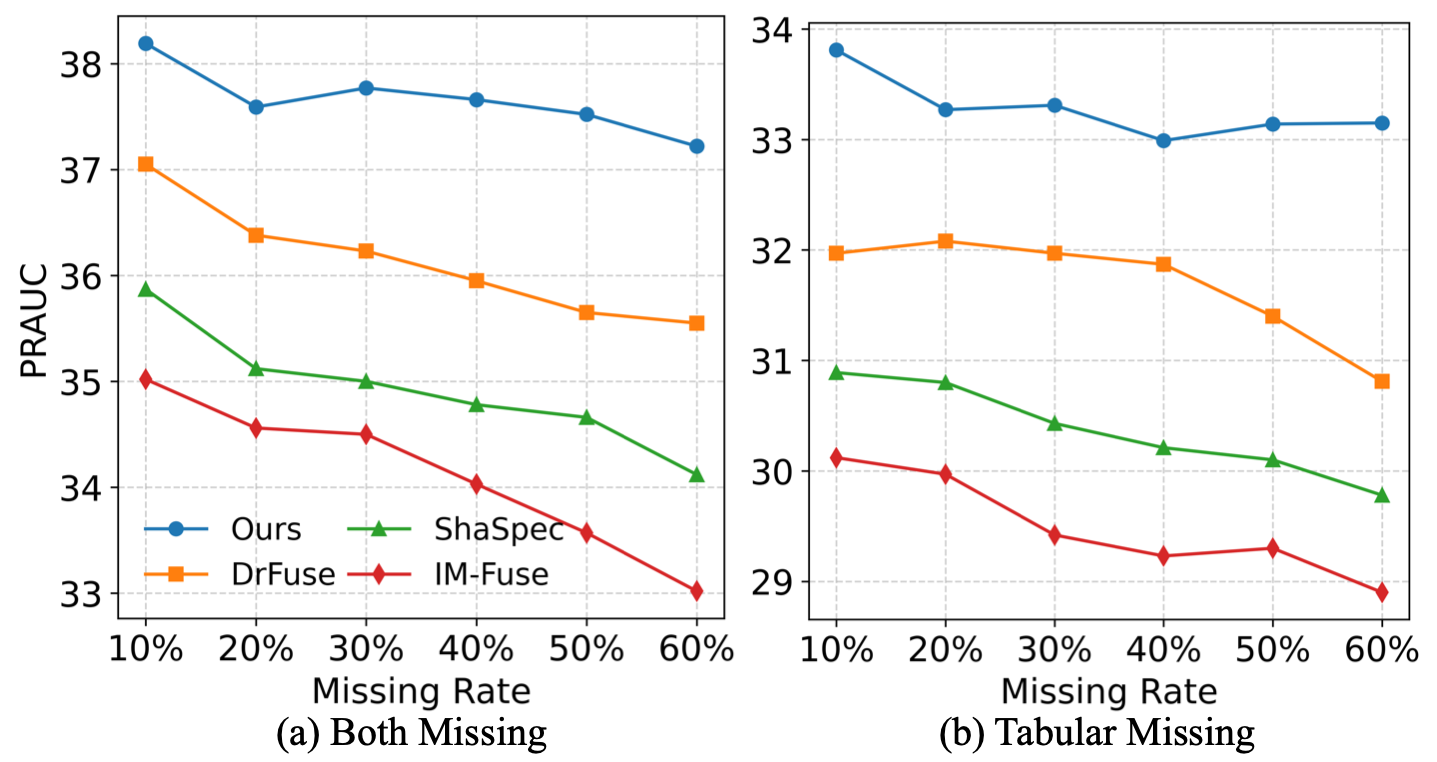}
   \caption{Generalization analysis on the MIMIC dataset under different training-time missing rates in terms of PRAUC. (c) Robustness analysis on the MIMIC dataset under different test-time missing rates in terms of PRAUC. 
   }
   \label{fig:general}
\end{figure}

\begin{figure}
  \centering
   \includegraphics[width=0.9\linewidth]{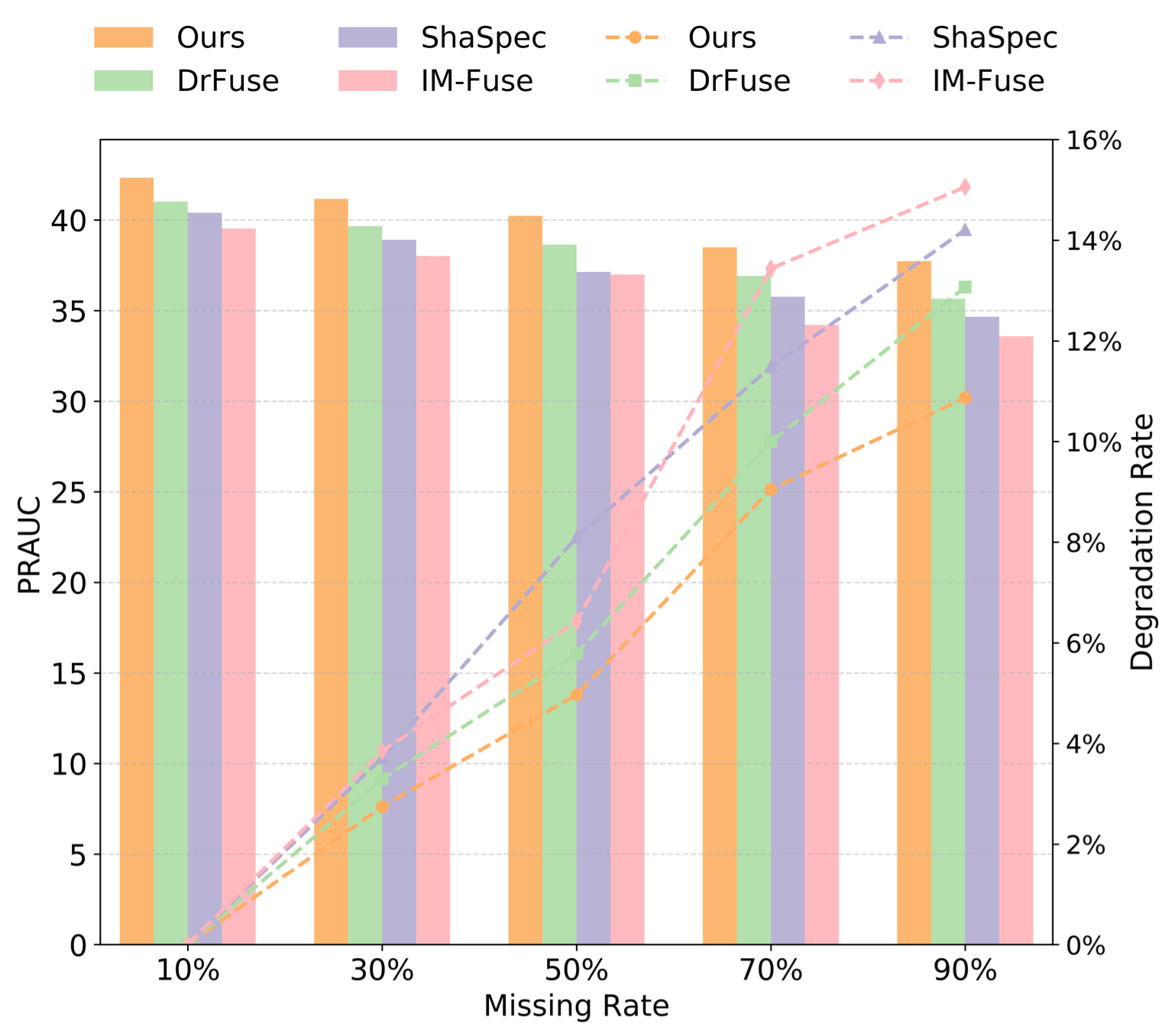}
   \caption{Robustness analysis on the MIMIC dataset under different test-time missing rates in terms of PRAUC. 
   }
   \label{fig:robust}
\end{figure}

\begin{table}[t]
\centering
\caption{Ablation study on three different components for MIMIC and DVM datasets under $50\%$ both-modality
missing. }
\label{tab:ablation}
        \begin{tabular}{ccc|c|c|c}
            \toprule
          \multicolumn{3}{c|}{Modules} & {MIMIC} & {ADNI} & {DVM} \\ 
          \cmidrule(lr){1-3} \cmidrule(lr){4-4} \cmidrule(lr){5-5} \cmidrule(lr){6-6} 
            SSPM & PFA & CMA & PRAUC (\%) & ACC (\%)  & ACC (\%)\\
            \midrule
            \xmark & \xmark & \xmark & 37.70  &74.14 & 89.67  \\
            \checkmark & \xmark & \xmark & 38.40 &74.77 & 91.32\\
            \checkmark & \checkmark & \xmark &39.50 &75.70 & 93.47  \\
            \checkmark & \checkmark & \checkmark & \textbf{40.22} &\textbf{76.48} & \textbf{94.50}  \\
            \bottomrule
        \end{tabular}
        \vspace{-5pt}
\end{table}

\noindent \textbf{Generalizability Analysis.}
Fig.~\ref{fig:general}(a) and (b) show the results on the MIMIC dataset under different training-time missing rates, evaluated on a 90\% missing test set. Specifically, Fig.~\ref{fig:general}(a) corresponds to the setting where both modalities are missing during training, while Fig.~\ref{fig:general}(b) represents the scenario where only the tabular modality is missing. DFPL consistently outperforms all baselines across all training-time missing rates, showing superior generalization ability in the presence of incomplete modalities. In particular, when the missing rate increases from 20\% to 60\%, our method maintains a relatively stable performance with only a slight decline, whereas competing methods degrade substantially. These results highlight DFPL’s strong generalizability, which can be attributed to its ability to effectively align cross-modal distributional and semantic cues.

\begin{table*}[t]
\centering
  \renewcommand{\arraystretch}{1} 
  \setlength\tabcolsep{10pt}
  \begin{minipage}[t]{0.3\textwidth}
    \renewcommand{\arraystretch}{0.8} 
   \caption{Ablations for the SSPM.}
\centering
  \label{tab:sspm}
  \begin{tabular}{cc}
    \toprule
    Configurations     & ACC (\%)\\
    \midrule
    w/o $\mathcal{L}_{\text{div}}$ &90.50 \\
    w/ $\mathcal{L}_{\text{div}}$ &\textbf{91.32}    \\
   \midrule
    $K$ = 3 &  89.25\\
    $K$ = 5 &  \textbf{91.32}\\
     $K$ = 10 & 90.34\\
     $K$ = 15 & 89.61\\
    \bottomrule
    \end{tabular}
    \end{minipage}
    \hfill
    \hspace{0.05\linewidth}
  \begin{minipage}[t]{0.3\textwidth}
  \caption{Ablations for the PFA.}
  \label{tab:pfa}
  \begin{tabular}{cc}
    \toprule
    Configurations     & ACC (\%)\\
    \hline
     w/ $\mathcal{L}_{\text{pali}}$ & 92.72\\
    w/ $\mathcal{L}_{\text{cali}}$ & 93.10\\
    w/ $\mathcal{L}_{\text{pali}}$ and  $\mathcal{L}_{\text{cali}}$  & \textbf{93.47}\\
    \hline
    w/ global-level CP & 92.23\\
    w/ prototype-level CP  & 92.52\\
    w/ both & \textbf{93.10}\\
     \bottomrule
  \end{tabular}
  \end{minipage}
  \hfill
  \begin{minipage}[t]{0.3\textwidth}
    \caption{Ablations for the CMA.}
    \centering
    \label{tab:cma}
    \begin{tabular}{cc}
     \toprule
      Configurations     & ACC (\%)\\
      \hline
      Concatenation & 93.47\\
      Addition & 92.50\\
      CMA & \textbf{94.50}\\
      \hline
      CMA (global) & 93.24\\
      CMA (prototype)& 93.67\\
      CMA (multi-scale) & \textbf{94.50}\\
      \bottomrule
    \end{tabular}
    \end{minipage}
\end{table*}

\noindent \textbf{Robustness Analysis.}
Fig.~\ref{fig:robust} presents the robustness results on MIMIC with models trained under 50\% both-modality missing and tested at different missing rates. DFPL consistently achieves the highest PRAUC across all test-time missing rates from 10\% to 90\%. As the missing rate increases, the performance of all methods inevitably declines, yet DFPL degrades more gracefully. This clearly showcases the robustness of our method to various missing patterns.

\subsection{Ablation Studies}

\noindent \textbf{Efficacy of Key Components.} We demonstrate the effectiveness of three
proposed components in DFPL: SSPM (Sec. \ref{sec:SSPM}), PFA (Sec. \ref{sec:PFA}), and CMA (Sec. \ref{sec:CMA}). To achieve this, we establish a baseline where only coarse-grained disentanglement and alignment are employed for shared–specific representation learning, followed by a simple concatenation of shared and specific representations for classification. We then progressively incorporate each proposed component into the baseline. Table \ref{tab:ablation} showcases that each component consistently improves performance on the MIMIC, ADNI, and DVM datasets. Incorporating SSPM yields a noticeable gain on the two datasets, verifying the benefit of fine-grained shared–specific prototype modeling. Adding PFA further boosts performance, highlighting its effectiveness in enhancing cross-modal semantic and distributional consistency. Finally, CMA brings additional improvements by adaptively aggregating multi-scale representations, leading to the best performance. Overall, compared with the baseline, DFPL achieves an absolute improvement of 2.52\% PRAUC on MIMIC, 2.34\% ACC on ADNI, and 4.83\% ACC on DVM, confirming the complementary effect of all three components.

\begin{figure}[t]
  \centering
   \includegraphics[width=1\linewidth]{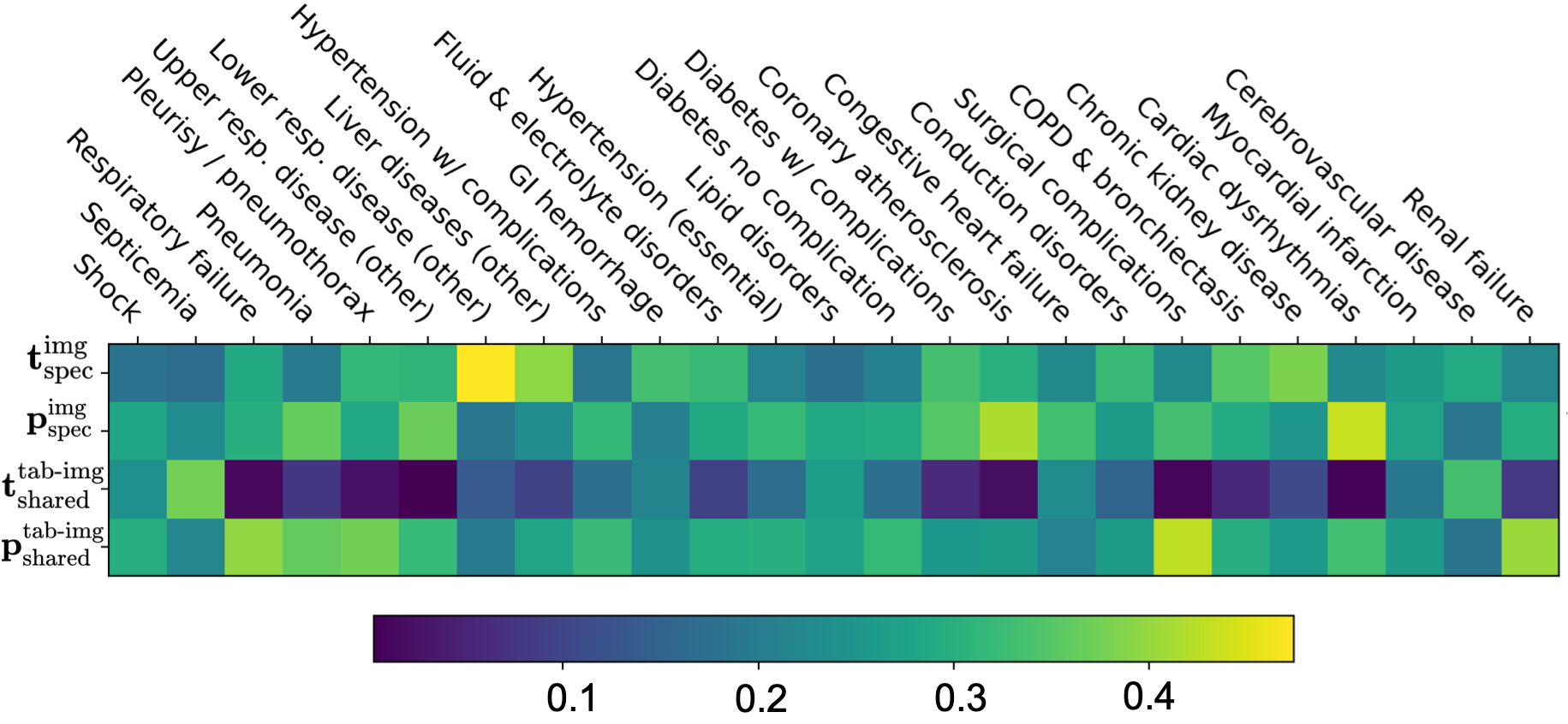}
   \caption{Visualization of cross-attention weights in CMA on the MIMIC dataset.}
   \label{fig:visatt}
\end{figure}

\noindent \textbf{Ablation Study on SSPM.} To study the effect of the diversity regularization loss and the number of prototypes in SSPM, we conduct experiments on the DVM dataset under $50\%$ missing rate, as shown in Table \ref{tab:sspm}. Incorporating the diversity loss $\mathcal{L}_{\text{div}}$ leads to a clear improvement, indicating its effectiveness in encouraging diverse prototype learning and preventing redundancy. We further vary the number of prototypes $K$ to analyze its impact. The best performance is achieved at $K=5$, whereas too few ($K=3$) or too many ($K=10,15$) prototypes lead to degradation. The main reason can be that a small $K$ limits representational capacity, while a large $K$ introduces redundancy and noise. Therefore, we adopt $K=5$ as the default setting in all subsequent experiments.

\noindent \textbf{Ablation Study on PFA.} In  Table \ref{tab:pfa}, we investigate the effect of the prototype-level and prototype-to-class alignment losses. Using either $\mathcal{L}_{\text{pali}}$ or $\mathcal{L}_{\text{cali}}$ alone leads to a performance gain, while combining them achieves the best result of 93.47\%, demonstrating their complementary roles. We then analyze the type of class prototypes (CP). Employing global- or prototype-level class prototypes alone results in 92.23\% and 92.52\% accuracy, respectively. When integrated through the proposed gating mechanism, the performance improves substantially, which can be attributed to the richer semantic cues provided by multi-level class prototypes for alignment.

\noindent \textbf{Ablation Study on CMA.} We further analyze the effect of different shared-specific fusion strategies, as summarized in Table \ref{tab:cma}. Compared with simple feature fusion methods such as concatenation or addition, CMA achieves a significant improvement. We also examine the impact of fusion at different scales. Compared with aggregation at the global and prototype levels, multi-scale aggregation yields improvements of 1.26\% and 0.83\%, respectively. These results show that CMA effectively leverages multi-scale information through class-aware aggregation, thereby producing more discriminative representations.

\begin{figure}[t]
  \centering
   \includegraphics[width=1\linewidth]{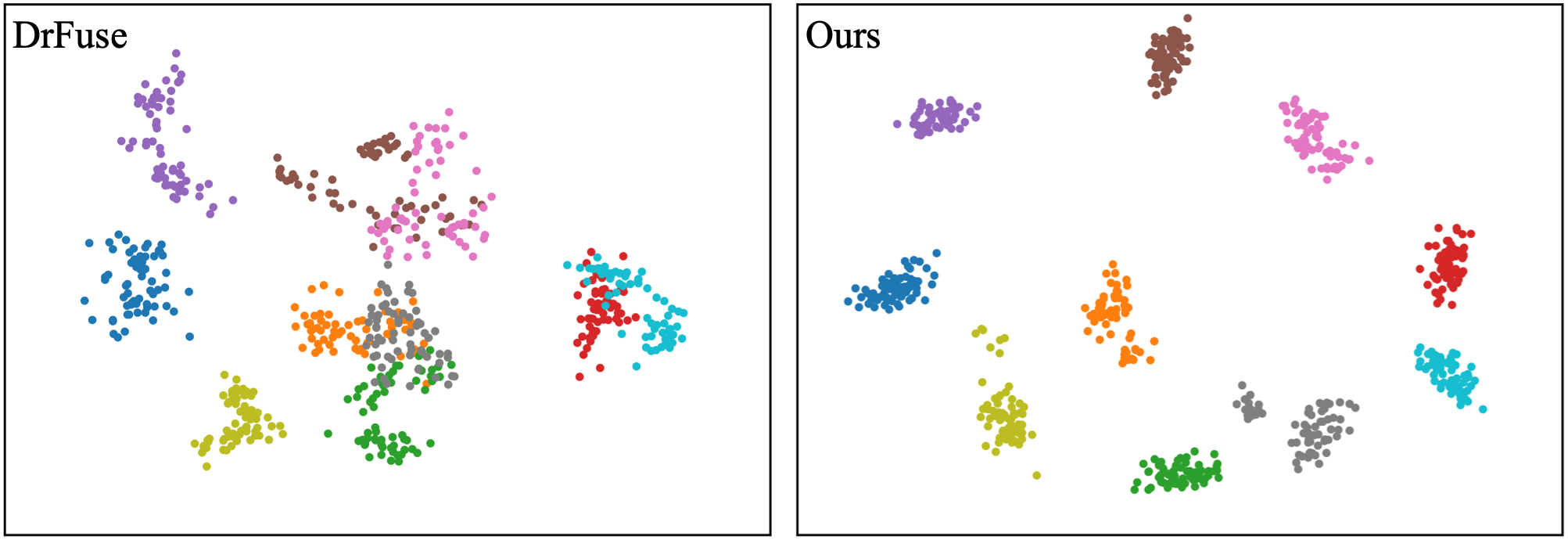}
   \caption{t-SNE visualization of class-related representations learned by DrFuse and our method. We show 10 classes on the DVM dataset in different colors.}
   \label{fig:tsne}
\end{figure}

\noindent \textbf{Qualitative results.} Fig.~\ref{fig:visatt} visualizes the cross-attention weights in CMA when the tabular modality is completely missing at test time. For each disease class, we average attention over samples with a prediction probability greater than 0.5. The attention patterns reveal that global- and prototype-level features provide complementary cues. When tabular data is absent, shared representations can offer valuable cues. Notably, for most disease classes, prototype-level shared features contribute more than global-level ones, highlighting the importance of prototype-level shared representations. We can also observe that different diseases rely on different feature levels. For example, the diagnosis of \emph{lower respiratory disease} appears to depend more on global-level image-specific features, whereas \emph{pneumonia} relies heavily on prototype-level image-specific and shared representations. These findings further validate the necessity of learning prototype-level representations.

To further examine the learned representations, we visualize class-related embeddings on the DVM validation set using t-SNE, as shown in Fig.~\ref{fig:tsne}. Compared with DrFuse (left), where the clusters remain relatively scattered and partially overlapping, our method (right) produces much clearer inter-class boundaries and more compact intra-class clusters. This indicates that DFPL is more effective at learning class-discriminative representations. Additional results are provided in \textbf{Supp. C}.

\section{Conclusion}

In this paper, we address the missing-modality challenge in heterogeneous image–tabular learning with a novel framework, DFPL. We first propose SSPM to extract compact and diverse shared and modality-specific prototypes from token-level features and disentangle them in the prototype space. We then introduce PFA to jointly align prototype-level distributions and prototype-to-class semantics, preserving fine-grained cross-modal distributional and semantic consistency. Finally, we design CMA to generate robust class-related information by adaptively fusing shared and specific information across global and prototype levels. Extensive experiments on three different kinds of image-tabular datasets demonstrate that DFPL achieves state-of-the-art performance under various missing-modality settings. Our current alignment strategy relies on paired modalities, leaving unpaired samples underutilized. Future work will investigate leveraging unpaired modalities for cross-modal knowledge transfer. Moreover, extending DFPL to other multimodal combinations beyond image–tabular data, such as image–text or audio–video, warrants further investigation.

\ifCLASSOPTIONcaptionsoff
  \newpage
\fi

\bibliographystyle{IEEEtran}
\bibliography{egbib}

\clearpage

\setcounter{table}{0}
\setcounter{figure}{0}
\setcounter{equation}{0}
\setcounter{page}{1}
\renewcommand\thefigure{S\arabic{figure}}
\renewcommand\thetable{S\arabic{table}}

\end{document}